\documentclass{article}

\usepackage[preprint]{neurips_2024}
\usepackage[pdftex]{graphicx}
\usepackage[dvipsnames]{xcolor}
\usepackage[utf8]{inputenc} 
\usepackage[T1]{fontenc}    
\usepackage{hyperref}       
\usepackage{url}            
\usepackage{booktabs}       
\usepackage{amsfonts}       
\usepackage{nicefrac}       
\usepackage{microtype}      
\usepackage{xcolor}         
\usepackage{float}
\usepackage[raster,most]{tcolorbox}
\usepackage{natbib}
\setcitestyle{numbers,square}
\usepackage{enumitem}
\usepackage{multirow}
\usepackage{multicol}

\title{{\protect\includegraphics[width=1em]{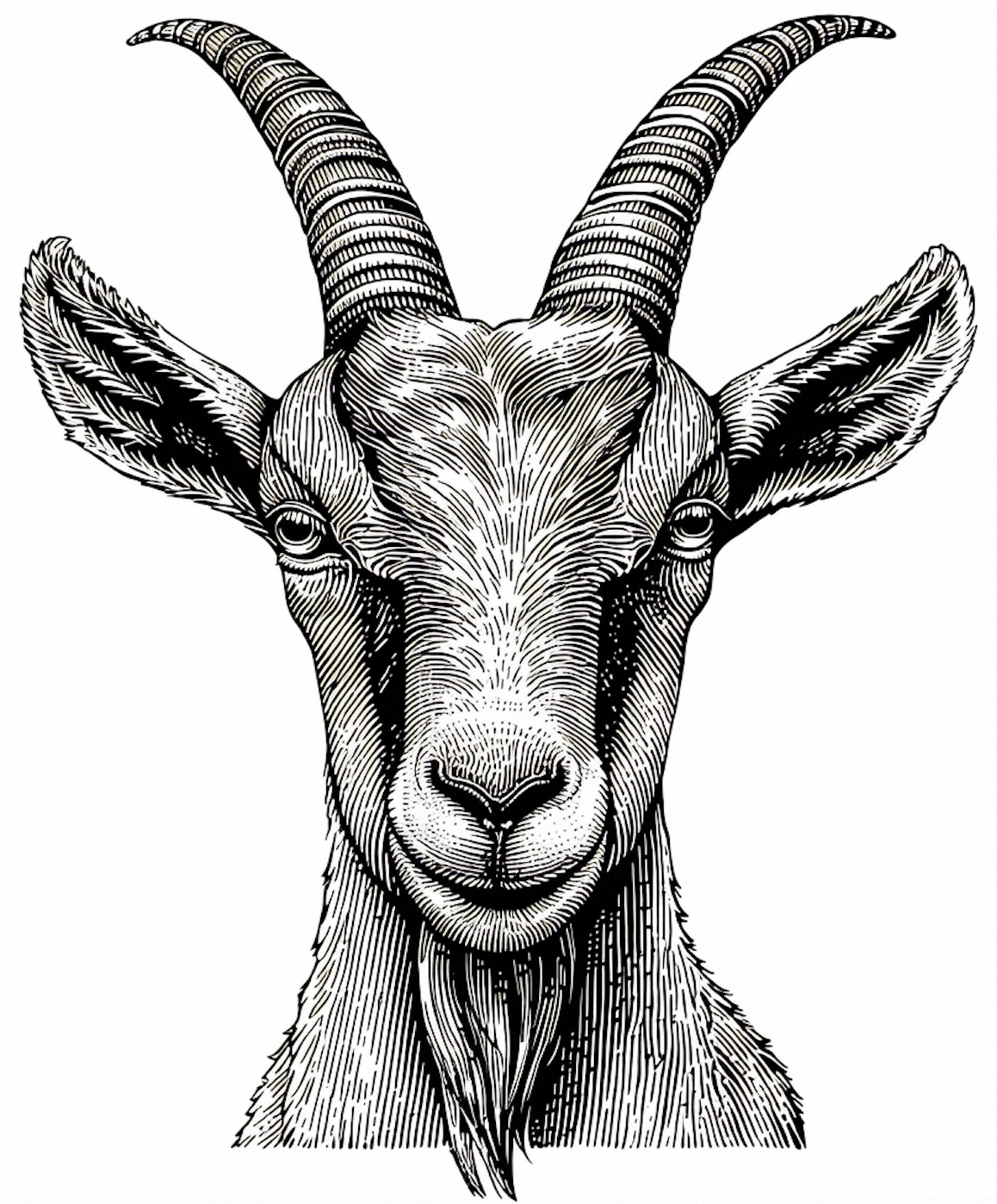}}GOAT-TTS: Expressive and Realistic \\Speech Generation via A Dual-Branch LLM}

\definecolor{gray}{HTML}{777777}
\definecolor{}{HTML}{00008B}

\author{
	Yaodong Song$^\dag$, Hongjie Chen$^\dag$, Jie Lian, Yuxin Zhang, Guangmin Xia, \\ \textbf{Zehan Li, Genliang Zhao, Jian Kang, Jie Li, Yongxiang Li, Xuelong Li$^*$} \\
	\\
	Institute of Artificial Intelligence (TeleAI), China Telecom, Beijing\\
	\\
	\texttt{xuelong\_li@chinatelecom.cn}\\
}

\begin{document}

	\maketitle
	
	\renewcommand{\thefootnote}{}
	\footnotetext{$^\dag$ Equal contribution.}
	\footnotetext{$^*$ Corresponding author.}
	\setcounter{footnote}{0}
	
	\begin{abstract}
		While large language models (LLMs) have revolutionized text-to-speech (TTS) synthesis through discrete tokenization paradigms, current architectures exhibit fundamental tensions between three critical dimensions: 1) irreversible loss of acoustic characteristics caused by quantization of speech prompts; 2) stringent dependence on precisely aligned prompt speech-text pairs that limit real-world deployment; and 3) catastrophic forgetting of the LLM's native text comprehension during optimization for speech token generation. 
		To address these challenges, we propose an LLM-based text-to-speech \textbf{G}eneration approach \textbf{O}ptimized via a novel dual-branch \textbf{A}rchi\textbf{T}ecture (GOAT-TTS). 
		Our framework introduces two key innovations: 
		(1) The modality-alignment branch combines a speech encoder and projector to capture continuous acoustic embeddings, enabling bidirectional correlation between paralinguistic features (language, timbre, emotion) and semantic text representations without transcript dependency; 
		(2) The speech-generation branch employs modular fine-tuning on top-k layers of an LLM for speech token prediction while freezing the bottom-n layers to preserve foundational linguistic knowledge. Moreover, multi-token prediction is introduced to support real-time streaming TTS synthesis. Experimental results demonstrate that our GOAT-TTS achieves performance comparable to state-of-the-art TTS models while validating the efficacy of synthesized dialect speech data.  
		
	\end{abstract}
	
	\section{Introduction}
	
	In the era of large language models (LLMs)\citep{qwen2,deepseek-v3,llama3}, the construction of end-to-end speech dialogue systems\cite{llama-omni, slam-omni, moshi, minmo} necessitating dialect comprehension and affective awareness typically demands the acquisition of large-scale dialect-specific and/or rich-emotion speech-text alignment datasets. 
	This requirement incurs significant resource expenditures due to the labor-intensive nature of multi-modal data curation. 
	However, given the relative abundance of textual data and the accelerated progress in LLM-driven speech synthesis architectures, we posit that this data scarcity challenge will be ameliorated through synthetic data augmentation.
	
	In particular, the field of speech synthesis has undergone transformative advances driven by breakthroughs in text-based language modeling paradigms. 
	Empirical studies across industry and academia have demonstrated that speech language models (SLMs) based on decoder-only transformers exhibit remarkable instruction-following capabilities, enabling granular control over vocal attributes such as speaker identity, environmental acoustics, affective prosody, and semantic coherence through speech prompt conditioning. 
	Pioneering works such as SPEAR-TTS\cite{SPEAR-TTS}, VALL-E\cite{vall-e}, and CosyVoice\cite{cosyvoice} have established foundational methodologies for few-shot/zero-shot speaker adaptation via autoregressive modeling of discrete speech tokens. 
	Recent architectural innovations exemplified by OuteTTS\cite{outetts}, CosyVoice2\cite{cosyvoice2}, and Llasa\cite{llasa} have further advanced the state-of-the-art by integrating pre-trained text language models as core architectural components.
	These approaches leverage the next-token prediction paradigm to speech tokens within the LLM's generative framework, achieving high-fidelity speech synthesis while maintaining linguistic coherence. 
	Collectively, these advancements underscore the cross-modal potential of large-scale pre-trained models, demonstrating their capacity to simultaneously enhance semantic understanding and high acoustic fidelity during the speech generation process. 
	
	However, our preliminary investigations reveal three intrinsic challenges in these methods.
	First, discrete tokenization of the speech prompt induces irreversible loss of acoustic characteristics, resulting in measurable degradation in terms of synthesized speech quality metrics including natural prosody and emotional expressiveness. 
	Second, the rigid association between audio signals and their corresponding transcripts in the prompt imposes significant constraints on the flexibility of real-world deployment. 
	Third, we observed a drastic deterioration of the LLM's native text comprehension capability during training, suggesting an inherent incompatibility between the preservation of pre-trained linguistic knowledge and the optimization of speech generation objectives.
	
	To address these challenges, we propose a novel TTS \textbf{G}eneration approach \textbf{O}ptimized by a dual-branch LLM \textbf{A}rchi\textbf{T}ecture (GOAT-TTS). 
	The framework comprises three core components, a shared pre-trained LLM backbone, a modality-alignment branch, and a speech-generation branch, achieving prompt-driven synthesis of high-quality speech through a two-stage training paradigm. 
	Specifically, the model employs a pretrained speech encoder to extract multidimensional acoustic features from a speech prompt. 
	Conditioned on the acoustic features, the LLM processes textual inputs to generate corresponding discrete speech tokens, which are subsequently mapped to mel-spectrograms via a flow-matching model.
	
	Our proposed GOAT-TTS framework differs from other existing LLM-based TTS systems in terms of architectural design and training methodologies.
	\begin{itemize}
		\item 
		First, through the modality-alignment branch, we extract continuous representations of the speech prompt by the combination of a speech encoder and a projector module. 
		Unlike conventional approaches that rely on discrete tokenization for speech prompt encoding, our method simultaneously captures rich paralinguistic features (e.g., timbral characteristics, prosodic contours, and affective nuances) and establishes bidirectional correlations between continuous acoustic embeddings and semantic text representations.    
		This deep alignment process equips the LLM backbone with the ability to directly interpret speech prompts through their continuous acoustic embeddings, eliminating the dependency on precise transcriptions of speech prompts during inference, which significantly enhances flexibility for industrial applications.
		\item  
		Second, to preserve the pre-trained LLM's intrinsic knowledge structure, the generation-branch adopts a layer-wise parameter-freezing strategy: only the top-k layers' parameters undergo task-specific fine-tuning while foundational layers remain frozen. This strategy ensures retention of the LLM's original contextual comprehension capabilities while enabling targeted optimization for speech synthesis objectives. Additionally, we introduce a multi-token prediction mechanism~\cite{mtp}, which reduces frequency discrepancies between the original LLM outputs and the fine-tuned model's speech token predictions. This design not only accelerates the inference speed but also inherently supports streaming text token input and streaming speech token output, offering general compatibility for real-world applications.
		
	\end{itemize}
	
	\section{Method}
	\label{sec:method}

\begin{figure}[t]
		\centering
		\includegraphics[width=1\linewidth]{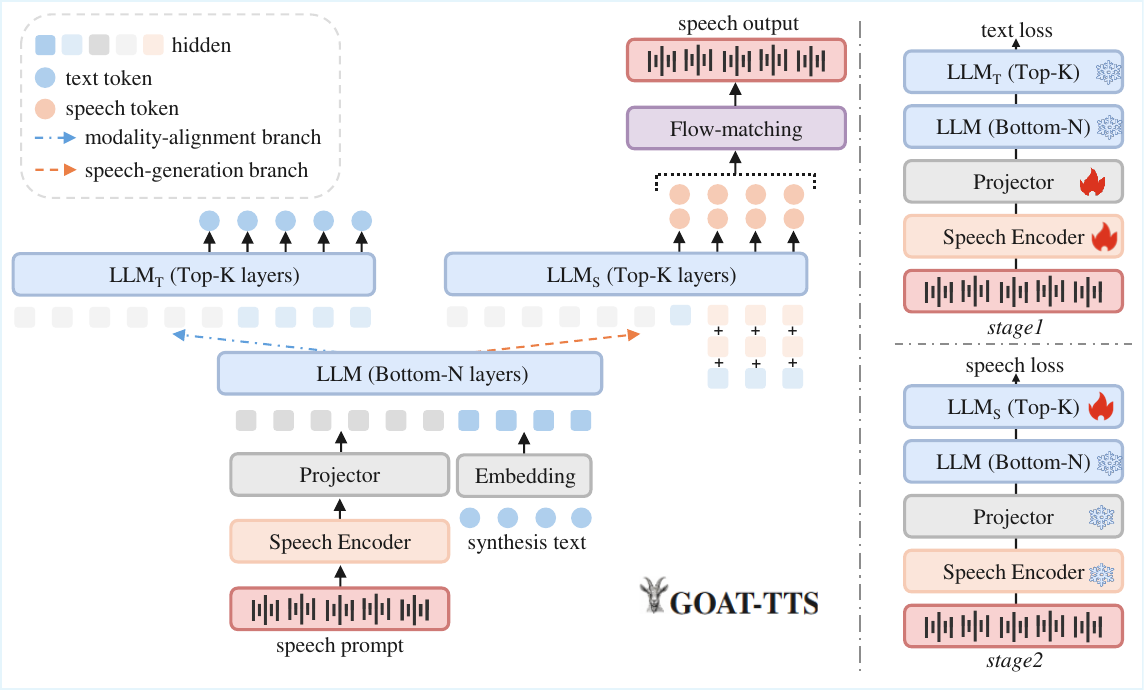} 
		\caption{Left: Model architecture of GOAT-TTS. Initially, $LLM_{S}$ and $LLM_{T}$ share weight parameters. Right: Illustration of the two-stage training strategy for GOAT-TTS.}
		\label{fig:YTTS}
	\end{figure}

	In this section, we detail the training methodology underpinning our proposed framework. As depicted in Figure \ref{fig:YTTS}, the framework employs a hierarchical processing pipeline. First, a pretrained Whisper-Small speech encoder is utilized to extract latent acoustic representations from input speech prompts. These latent features are then mapped to LLM-compatible embeddings through a projector module, which incorporates a 3-layer CNN-based architecture followed by a linear transformation layer. Subsequently, the LLM generates speech tokens in an iterative fashion, conditioned jointly on both the adapted acoustic embeddings and the provided textual input. Finally, the discrete speech tokens are converted into mel-spectrogram representations via a flow-matching model, which facilitates subsequent waveform synthesis through probabilistic density estimation.
	
	The training regimen for the complete framework is structured into two sequential stages. Stage I centers on modality-alignment training, which establishes cross-modal semantic coherence between acoustic and linguistic features by optimizing the speech encoder and projector module. Stage II subsequently transitions to speech-generation training, where the full pipeline is fine-tuned to refine the end-to-end speech synthesis capability. During this stage, the flow-matching model is further optimized to enhance waveform fidelity by minimizing the discrepancy between predicted and target mel-spectrograms. This phased approach ensures robust modality alignment while enabling high-quality speech generation through iterative parameter optimization.
	
	\subsection{Modality-Alignment Training}
	Inspired by previous work on modality alignment \cite{BLSP, AudioChatLlaMA}, our framework adopts speech-text continuation pairs as the core training data for speech-text alignment. In this stage, the LLM takes speech-derived query embeddings as input and generates text continuations that are generated by the LLM when it consumes the corresponding speech transcripts. Two complementary strategies are proposed to construct a sufficiently large and diverse dataset of such pairs.
	We first exploit transcripts from large-scale automatic speech recognition (ASR) corpora, which encompass Mandarin, English, and multiple dialects. These transcripts are extended by prompting the LLM to generate contextually coherent continuation text. This approach leverages the acoustic-semantic correlations embedded in ASR corpora.
	Second, we employ TTS to convert semantically coherent sentences into speech audio, while the LLM simultaneously generates continuation text for these sentences. These two processes ensure that the training data maintains both broad acoustic and semantic diversity.
	To encode modality-specific characteristics, we prepend natural language descriptors (e.g., dialect specifications, emotional cues) to the input text during continuation generation. This explicit contextual guidance enables the model to learn modality-dependent representations for dialectal variations and affective expressions.
	This structured approach addresses the challenge of data scarcity in cross-modal alignment.
	
	The training process is divided into two steps. Step I focuses on training with Mandarin and English data to establish a robust semantic parsing foundation for the projector module. This step mitigates training instability caused by excessive gradient variance during early optimization.
	Step II further optimizes the encoder and the projector on a carefully balanced dataset in terms of languages/dialects and emotions.
	The LLM’s parameters remain frozen in both Step I and Step II.
	This phased approach integrates multilingual, dialectal information into a unified semantic embedding space, ensuring cross-modal consistency and generalization across speech and text modalities.
	
	\subsection{Speech-Generation Training}
	The speech-generation training leverages large-scale real-world datasets ($\sim$150k hours) organized as quadruple, <text query, speech query, text response, speech response>, where the speech query and speech response adhere to consistent speaker-specific conversational characteristics. Implemented through supervised fine-tuning (SFT), this stage utilizes:
	\begin{itemize}[label=$\bullet$, nosep, leftmargin=*, itemsep=0pt, topsep=0pt]
		\item Input: Concatenated hidden from speech query or text query, text response and speech response tokens.
		\item Target: Ground-truth speech response tokens.
		\item Objective: Cross-entropy loss minimization on speech tokens prediction.
	\end{itemize}
	The training process consists of two steps, as illustrated in Table \ref{tab:speech-training}. In Step I, we select <text query, text response, speech response> pairs as inputs for Mandarin and English data, while adopting <speech query, text response, speech response> pairs for dialect data to jointly train the cold-start model. This approach is designed to achieve dual objectives: enabling dialect adaptation while maintaining stable generation capabilities in both Mandarin and English. In Step II, building upon the cold-start model, we continue training using <speech query, text response, speech response> pairs as inputs to develop prompt-driven speech generation capabilities. Our experiments demonstrate that compared with directly using <speech query, text response, speech response> pairs as inputs, this two-step training strategy yields more stable model performance, particularly in cross-lingual generation scenarios.
	
	\begin{table*}[h]
		\centering
		\caption{The data configuration for GOAT-TTS speech-generation training.}
		\setlength\tabcolsep{6pt}
		\scalebox{0.90}{
			\begin{tabular}{lcccccc}
				\toprule
				\multicolumn{1}{c}{\textbf{Training}} & \multicolumn{1}{c}{\textbf{Language}} & \multicolumn{1}{c}{\textbf{Data Structure}}\\
				\midrule
				\multirow{2}{*}{\textbf{Step I}} & \multicolumn{1}{c}{Mandarin and English}  & \multicolumn{1}{c}{<text query, text response, speech response>}\\
				\cmidrule(r){2-2} \cmidrule(r){3-3}
				& Dialects & \multicolumn{1}{c}{<speech query, text response, speech response>}   \\
				\midrule
				\multirow{1}{*}{\textbf{Step II}} & \multicolumn{1}{c}{Mandarin, English and Dialects}  & \multicolumn{1}{c}{<speech query, text response, speech response>}\\
				\bottomrule
		\end{tabular}}
		\label{tab:speech-training}
	\end{table*}
	
	For parameter optimization, we adopt a modular training strategy where the speech encoder, projector, and foundational layers of the LLM are frozen to preserve pre-trained acoustic and linguistic representations. Meanwhile, the top-k semantic-to-speech token mapping layers are fine-tuned to adapt cross-modal interactions. In our implementation, we set $N=\lfloor M / 2 \rfloor$ and $K=M-N$, where $M$ is the total number of the LLM layers. We resort the further investigation of other possible settings to future work. Furthermore, a critical innovation is the multi-token prediction mechanism, which propagates temporal coherence by concatenating the current-step output token embeddings with next-step input features within the frozen upper network parameters. This fusion of sequential contextual information generates a temporally coherent latent representation, which guides the iterative generation of subsequent speech tokens. 
	
	
	During inference, the system operates as follows: The speech query (input speech prompt) and text response (target synthesis text) are jointly processed through the pipeline. The generated speech tokens are then converted into the final synthesized speech output via the flow-matching model, which performs waveform synthesis by estimating the conditional probability distribution of mel-spectrograms. This design ensures that both modalities are dynamically integrated while maintaining computational efficiency through frozen module utilization.
	
	
	\section{Experiments}

	\subsection{Comparison Results with Baselines}
	
	We first evaluate our GOAT-TTS model on  the commonly-used test sets: SEED \emph{test-zh}, \emph{test-en} and \emph{test-hard} ~\cite{Seed-TTS}, compared it with several top popular open-source zero-shot TTS models, such as FireRedTTS~\cite{fireredtts}, MaskGCT~\cite{maskgct}, E2 TTS~\cite{E2-TTS}, F5-TTS~\cite{F5-TTS}, Llasa~\cite{llasa}, CosyVoice~\cite{cosyvoice} and CosyVoice 2(streaming and non-streaming)~\cite{cosyvoice2}. The objective evalution results for GOAT-TTS and baseline models are presented in Table~\ref{tab:seedttseval}. From the table, compared to non-streaming inference results, our model achieves comparable performance. GOAT-TTS ranks third on the \emph{test-zh}, trailing only CosyVoice2 and FireRedTTS. Analysis of generated audio reveals several samples (from the same prompt) that are erroneously synthesized in dialect—while pronunciation accuracy is maintained, the Paraformerzh model exhibits suboptimal recognition on these samples, leading to a high Character Error Rate (CER). On the \emph{test-hard} and \emph{test-en}, our model ranks second and third respectively, further demonstrating its stability and robustness for diverse scenario prompts. Notably, compared to streaming inference results, GOAT-TTS comprehensively outperforms CosyVoice2-S in both \emph{test-en} and \emph{test-hard}, validating the significant advantages of our native streaming inference architecture.

	\begin{table*}[h]
		\centering
		\caption{Results of Our and recent TTS models on the SEED test sets\cite{Seed-TTS}. $\dagger$ denotes close-sourced models. For WER, we employ Whisper-large-v3 \cite{whisper-large-v3} and Paraformerzh \cite{paraformer} as the automatic speech recognition (ASR) engines for English and Mandarin, respectively.}
		{
			\begin{tabular}{lcccccc}
				\toprule
				\multirow{2}{*}{\textbf{Model}} & \multicolumn{1}{c}{\textbf{\emph{test-zh}}} & \multicolumn{1}{c}{\textbf{\emph{test-en}}} & \multicolumn{1}{c}{\textbf{\emph{test-hard}}} \\
				\cmidrule(r){2-2} \cmidrule(r){3-3} \cmidrule(r){4-4}
				& \textbf{CER (\%)~$\downarrow$} & \textbf{WER (\%)~$\downarrow$} & \textbf{WER (\%)~$\downarrow$}   \\
				\midrule
				\textbf{Human} & 1.26  & 2.14   & -  \\
				\midrule
				\textbf{FireRedTTS}~\cite{fireredtts} & 1.51   & 3.82  & 17.45 \\
				\textbf{MaskGCT}~\cite{maskgct} & 2.27  & 2.62 & 10.27 \\
				\textbf{E2 TTS (32 NFE)}$^\dagger$~\cite{E2-TTS} & 1.97  & 2.19 & - \\
				\textbf{F5-TTS (32 NFE)}~\cite{F5-TTS} & 1.56  & 1.83 & 8.67 \\
				\textbf{Llasa-8B}~\cite{llasa} & 1.59 & 2.97 & 11.09  \\
				\textbf{CosyVoice}~\cite{cosyvoice} & 3.63  & 4.29 & 11.75 \\
				\textbf{CosyVoice 2}~\cite{cosyvoice2} & 1.45  & 2.57 &  6.83  \\
				\textbf{CosyVoice 2-S}~\cite{cosyvoice2} & 1.45  & 2.38  & 8.08  \\
				\midrule
				\textbf{Our} & 1.53  & 2.24  & 7.83  \\
				\bottomrule
		\end{tabular}}
		\label{tab:seedttseval}
	\end{table*}

\subsection{Dialect ASR Evaluation}\label{sec:Dialect-ASR-Evaluation}
Through modal-alignment training and speech-generation training, our GOAT-TTS model can generate high-quality dialect-specific speech outputs when solely driven by prompts, which provides valuable training data for ASR models.
We also validate the efficacy of our approach in synthesizing dialect speech data by conducting dialect Automatic Speech Recognition (ASR) experiments, utilizing both open-source and internal dialect corpus.

\subsubsection{Dialect ASR Evaluation on Open-Source Corpus}\label{sec:Opensource-ASR}
Zhongyuan and Southwestern dialects from the open-source KeSpeech dataset~\cite{kespeech} are selected as the target dialects. Speech prompts are randomly sampled from the training sets of these dialects, while all transcripts in the training sets are synthesized using both our GOAT-TTS and CosyVoice2. Zipformer-based automatic speech recognition (ASR) models~\cite{zipformer}, initialized with parameters pre-trained on 10,000-hour Mandarin data, are trained separately on synthetic and raw data. 

\begin{table}[h]
	\centering{%
		\caption{ASR Word Error Rate (WER) results on Zhongyuan and Southwestern dialect test sets. GT denotes the model trained on raw data, and Baseline denotes the initial ASR model.}\label{tab:kespeech-wer}
		\begin{tabular}{lcccccc}
			\toprule
			\multirow{2}{*}{\textbf{Method}} & 
			\multicolumn{2}{c}{\textbf{ASR Word Error Rate(\%)}}                                 \\ 
			\cmidrule(r){2-3}
			& \textbf{Zhongyuan}     & \textbf{Southwestern}    \\ 
			\midrule
			\textbf{GT} & 7.54 & 9.58 \\ 
			\midrule
			\textbf{CosyVoice2}  & 20.01  & 19.52  \\
			\textbf{Baseline} & 38.56 & 38.25 \\ 
			\midrule
			\textbf{Our} & 15.82 & 13.0 \\ 
			\bottomrule
		\end{tabular}
	}
	
\end{table}
The ASR performance of raw data and synthesized data for the two dialects is summarized in Table~\ref{tab:kespeech-wer}.
The table demonstrates that our method outperforms CosyVoice2, with notable reductions in WER of 21.0\% and 33.4\% for Zhongyuan and Southwestern dialects, respectively. While the experimental results still exhibit a quantifiable gap relative to Ground Truth (GT) benchmarks, they consistently outperform baseline ASR models with WER improvements exceeding 59\%. The observed performance disparity between our method and GT may be attributed to dialect-specific discrepancies in training data composition. Specifically, while our TTS training corpus incorporates samples from Henan and Sichuan dialects, it may not sufficiently capture the phonological and lexical diversity inherent in broader Zhongyuan and Southwestern Mandarin dialect groups. This gap underscores the need for future work focused on targeted data augmentation and model adaptability to address regional linguistic variations better.

\subsubsection{Dialect ASR Evaluation on Internal Corpus}
Employing the same ASR model configuration and training pipeline as that of Section~\ref{sec:Opensource-ASR} , we validate the effectiveness of our approach across five internal dialects. Table~\ref{tab:internal-wer} details the experimental results. As shown in Table~\ref{tab:internal-wer}, our method significantly outperforms the baseline model, particularly in Cantonese, Shanghai and Sichuan dialects, where the WER decreases by over 70\%. Compared to the GT model, our approach achieves comparable performance, reaching about 82.5\% $\sim$ 88.9\% of its performance in Cantonese, Northeastern, Henan, and Shanghai dialects. These findings underscore the robustness of our synthesized speech data for ASR tasks, as they not only narrow the gap with GT performance but also demonstrate generalizability across diverse dialectal variations.
\begin{table}[h]
	\centering{%
		\caption{ASR Word Error Rate (WER) results on 5 target dialect test sets.}
		\label{tab:internal-wer}
		\begin{tabular}{lcccccc}
			\toprule
			\multirow{2}{*}{\textbf{Method}} & 
			\multicolumn{5}{c}{\textbf{ASR Word Error Rate(\%)}}                                 \\ 
			\cmidrule(r){2-6}
			& \textbf{Cantonese}     & \textbf{Northeastern} & \textbf{Henan} 
			& \textbf{Shanghai} & \textbf{Sichuan} \\ 
			\midrule
			\textbf{Baseline} & 89.54 & 13.63 & 61.91 & 76.81 & 41.53 \\ 
			\textbf{GT} & 23.83 & 5.55 & 21.16 & 15.6 & 11.5 \\ 
			\midrule
			\textbf{Our} & 26.48 & 6.52 & 24.19 & 17.34 & 14.36 \\ 
			\bottomrule
		\end{tabular}
	}
\end{table}

The consistent WER reductions—exceeding 70\% in high-variance dialects like Cantonese—highlight the model’s capacity to capture nuanced phonetic and prosodic features critical for dialectal ASR. While the remaining gap relative to GT performance (e.g., relative 24\% in Southwestern) suggests room for improvement, this disparity likely stems from the limited diversity in our training corpus. Specifically, the Sichuan dialect corpus for our TTS model does not fully encapsulate the broader phonological and lexical heterogeneity within Southwestern Mandarin.

These results validate the practical utility of our framework for real-world ASR applications, particularly in scenarios where dialectal diversity poses challenges for traditional systems. By generating speech data that aligns closely with natural dialectal characteristics, our method provides a scalable solution for improving ASR robustness without requiring extensive dialect-specific training data.

\section{Conclusion}
This report presents GOAT-TTS, a dual-branch generative architecture that synergizes large language model (LLM) capabilities with native streaming speech synthesis. By integrating continuous acoustic embeddings via a modality-alignment branch and preserving foundational linguistic knowledge through layer-wise parameter freezing, our framework achieves high-fidelity speech generation. The proposed two-stage training strategy and multi-token prediction mechanism further address the inherent incompatibility between speech generation optimization and LLM knowledge retention, enabling efficient streaming synthesis compatible with real-world applications. Crucially, our methodology validates the feasibility of leveraging synthetic data augmentation to mitigate dialectal data scarcity, thereby reducing dependency on labor-intensive datasets.  

\section{Future Work}
The experiment preliminarily validates that synthesized dialect speech data can be used to improve the performance of ASR models. We will further explore the application of synthesized speech data in more scenarios, such as enhancing the performance of TTS models and end-to-end speech-to-speech models.

Through modal-alignment training and speech-generation training, we believe that the GOAT-TTS framework has acquired preliminary end-to-end speech-to-speech capabilities. We will further explore its potential and applications in developing end-to-end speech-to-speech assistant models.  

\bibliographystyle{IEEEtran}
\bibliography{ref}
\end{document}